\documentclass[10pt,twocolumn,letterpaper]{article}

\usepackage{cvpr}              %

\definecolor{cvprblue}{rgb}{0.21,0.49,0.74}
\usepackage[pagebackref,breaklinks,colorlinks,allcolors=cvprblue]{hyperref}

\def\paperID{3684} %
\def\confName{CVPR}
\def\confYear{2026}
\usepackage{colortbl}
\usepackage{xcolor}
\definecolor{yellow}{rgb}{1, 1, 0.7}
\definecolor{orange}{rgb}{1, 0.85, 0.7}
\definecolor{red}{rgb}{1, 0.7, 0.7}
\definecolor{normalred}{rgb}{1, 0, 0}

\begin{document}

\title{4C4D: 4 Camera 4D Gaussian Splatting}
 
\author{
Junsheng Zhou$^{1*}$ \quad Zhifan Yang$^{1*}$ \quad Liang Han$^{1}$ \quad Wenyuan Zhang$^{1}$ \\[2mm]
Kanle Shi$^2$ \quad Shenkun Xu$^{2}$ \quad Yu-Shen Liu$^{1\dag}$ \\[4mm]
$^1$School of Software, Tsinghua University \qquad $^2$Kuaishou Technology \\[2mm]
{\tt\small \{zhou-js24,yangzf22,hanl23,zhangwen21\}@mails.tsinghua.edu.cn}\\ 
{\tt\small \{shikanle,xushenkun\}@kuaishou.com,liuyushen@tsinghua.edu.cn}
}

\twocolumn[{%
\renewcommand\twocolumn[1][]{#1}%
\maketitle
\begin{center}
\centering
\captionsetup{type=figure}
\vspace{-6mm}
\includegraphics[width=1\linewidth]{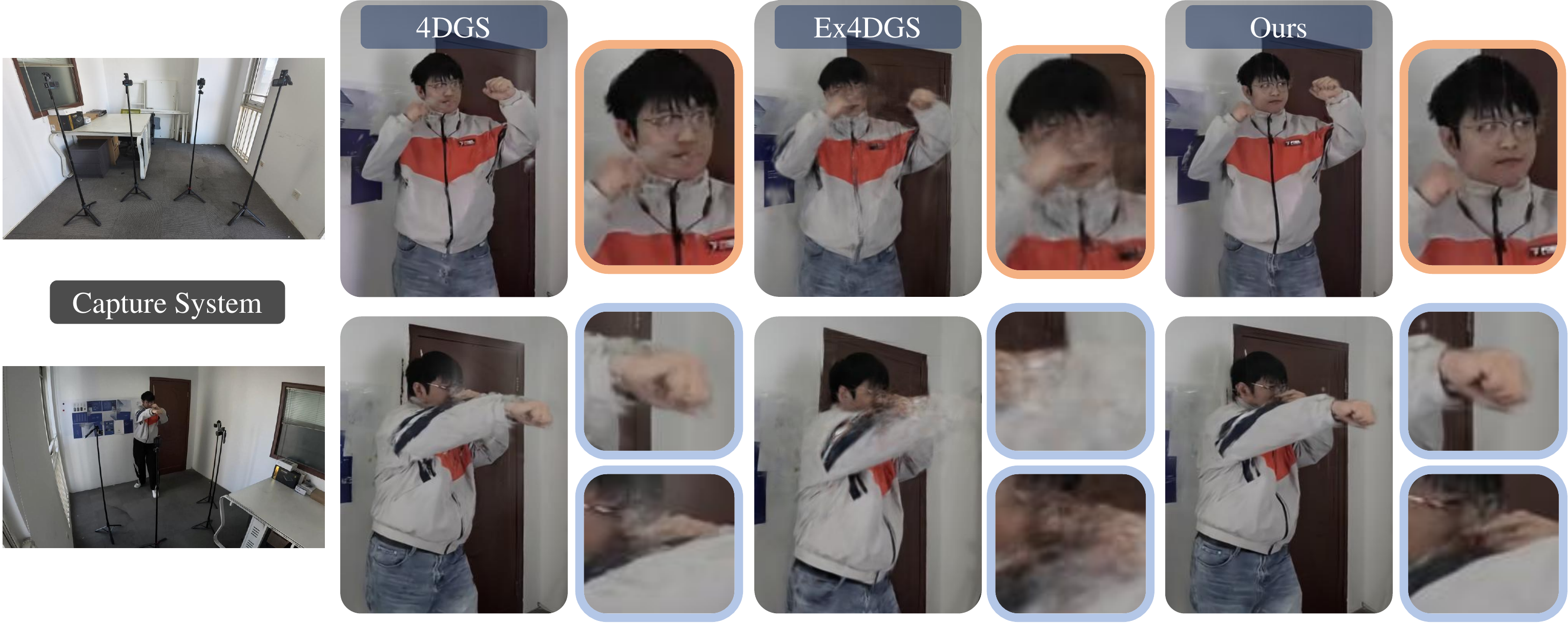}
\vspace{-6mm}
\captionof{figure}{\textbf{Left:} Our capture system records dynamic scenes using only four cameras. \textbf{Right:} Dynamic novel views rendered from the reconstructed 4D representations produced by Ex4DGS, 4DGS, and our method.
}
\label{fig:teaser}
\end{center}%
}]

\if TT\insert\footins{\footnotesize{
*Equal contribution. $\dag$ Corresponding author.}}\fi

\begin{abstract}
This paper tackles the challenge of recovering 4D dynamic scenes from videos captured by as few as four portable cameras. Learning to model scene dynamics for temporally consistent novel-view rendering is a foundational task in computer graphics, where previous works often require dense multi-view captures using camera arrays of dozens or even hundreds of views. We propose \textbf{4C4D}, a novel framework that enables high-fidelity 4D Gaussian Splatting from video captures of extremely sparse cameras. Our key insight lies that the geometric learning under sparse settings is substantially more difficult than modeling appearance. Driven by this observation, we introduce a Neural Decaying Function on Gaussian opacities for enhancing the geometric modeling capability of 4D Gaussians. This design mitigates the inherent imbalance between geometry and appearance modeling in 4DGS by encouraging the 4DGS gradients to focus more on geometric learning. Extensive experiments across sparse-view datasets with varying camera overlaps show that 4C4D achieves superior performance over prior art. Project page at: \url{https://junshengzhou.github.io/4C4D}.
\end{abstract}

\section{Introduction}

Novel view synthesis of dynamic scenes is a vital task towards achieving 4D intelligence in various applications, including virtual reality, game design and film production. Prior works \cite{yang2023real,wu20244d} reconstruct dynamic scenes using complex systems of dense camera arrays with hundreds of synchronized devices, which confines 4D modeling to research labs or large companies rather than everyday users. In this paper, we address the extremely challenging problem of reconstructing 4D dynamic scenes from sparse camera setups. This technique provides a promising pathway to bring 4D intelligence into daily life, enabling high-quality dynamic scene reconstruction using only a few mobile phones or portable cameras such as GoPros.

Recent advances \cite{kerbl20233dgs,mildenhall2020nerf} in novel view synthesis of 3D static scenes have demonstrated remarkable performance in both rendering quality and efficiency. 3D Gaussian Splatting (3DGS) \cite{kerbl20233dgs} represents scenes with explicit graphic primitives, enabling real-time photorealistic rendering through rasterization. Extending beyond static observations, 4D Gaussian Splatting \cite{yang2023real} and deformation-based 3DGS approaches \cite{wu20244d,yang2024deformable} further advance dynamic scene representation by integrating temporal modeling into Gaussian Splatting. Nevertheless, these methods remain applicable primarily to 4D reconstruction from videos captured by dense camera arrays, which limits their usability in sparse or consumer-level settings.

Through a deep dive into the optimization in 4D Gaussian Splatting under sparse-view settings, we observe that the failure of existing 4D frameworks can largely be attributed to a biased optimization process. Specifically, fitting dynamic appearance remains relatively easy for 4DGS even with limited views, whereas recovering accurate and noise-free geometry is significantly more challenging due to insufficient spatial supervision. The current Gaussian formulation fails to properly balance appearance and geometry learning, resulting in biased optimization that overfits to the easier appearance cues while neglecting the underlying geometric consistency.
 
Driven by this observation, we propose 4C4D, a novel 4D Gaussian Splatting framework for reconstructing high-fidelity dynamic scenes from as few as four camera views. 4C4D addresses the inherent imbalance between geometry and appearance learning in sparse-view 4DGS by introducing an adaptive learning scheme that grants more flexibility to geometric optimization, leading to a more balanced overall training process. To achieve this, we propose the Neural Decaying Function, a novel mechanism implemented with neural networks that adaptively controls the decay of Gaussian opacities. The Neural Decaying Function takes key attributes of each 4D Gaussian primitive  (e.g. center, opacity, rotation) as input and predicts a decay factor that neurally modulates Gaussian opacity. By introducing neural control over the opacity, which plays a crucial role in guiding 4DGS geometric learning, 4C4D directs the optimization gradients to focus more effectively on the learning of scene geometry, thereby achieving a better balance between geometric fidelity and appearance modeling.

Simply adopting the Neural Decaying Function across all 4D Gaussians leads to suboptimal training. The reason is that the gradients only exist in the visible regions of 4DGS that contribute to Gaussian rendering, while applying the same decay function to the opacities of invisible regions distorts the optimization process and leads to performance degradation. To address this issue, we introduce a visibility detection strategy that operates across both spatial and temporal domains to identify visible 4D Gaussians at each view and timestep. We then apply separate decay strategies for visible and invisible regions, ensuring that the Gaussian gradients correctly update the Neural Decaying Function during optimization. We evaluate 4C4D on dynamic scene reconstruction from sparse-view inputs across multiple datasets, including Neural3DV, ENeRF-Outdoor, Mobile-Stage and our self-captured Dyn4Cam, which cover both large and small camera overlaps. Experimental results demonstrate that 4C4D achieves superior and temporally consistent novel-view synthesis compared to existing state-of-the-art methods. Our main contributions can be summarized as follows.  

\begin{itemize}
\item We propose 4C4D, a novel 4D Gaussian Splatting framework that reconstructs high-fidelity dynamic scenes from sparse camera views as few as four, enabling temporally consistent and flexible novel-view rendering across all time steps.

\item We introduce the Neural Decaying Function, which adaptively guides Gaussian geometric learning by encouraging 4DGS gradient to focus more on Gaussian opacities. We further design separate decay strategies for visible and invisible regions to ensure stable optimization.

\item We evaluate 4C4D on dynamic scene modeling across sparse-view datasets with varying camera overlaps, and show our superiority over the state-of-the-art methods. 

\end{itemize}

\section{Related Work}

\subsection{Novel View Synthesis}

Novel view synthesis \cite{mildenhall2020nerf,barron2022mip,chen2022tensorf,kerbl20233dgs} for static 3D scenes is a long-standing task in 3D computer vision that has drawn significant attention since the introduction of Neural Radiance Fields (NeRF) \cite{mildenhall2020nerf}. NeRF achieves high-quality novel view synthesis by performing volume rendering over view-dependent attributes predicted by a neural network–based function. Subsequent works have extended NeRF to various scenarios, including sparse-view inputs \cite{truong2023sparf,huang2023neusurf,shi2024zerorf}, large-scale scenes~\cite{tancik2022block,turki2022mega}, surface reconstruction~\cite{wang2021neus,wang2022improved,oechsle2021unisurf}, and fast learning~\cite{muller2022instant,chen2022tensorf,fridovich2022plenoxels,zhou2024fast}.

Nevertheless, the NeRF family of methods suffers from limited rendering efficiency, which hinders their deployment in real-world applications requiring real-time rendering. Recently, 3D Gaussian Splatting (3DGS)~\cite{kerbl20233dgs} has been proposed, advancing novel view synthesis by enabling real-time rendering through rasterization of explicit Gaussian primitives. Beyond static reconstruction, 3DGS has also been explored for 3D generation~\cite{zhou2024diffgs,he2024gvgen,tang2023dreamgaussian,zhang2025gap,zhou2023uni3d}, reconstruction~\cite{chen2024mvsplat,zhu2024fsgs,han2024binocular}, and material modeling~\cite{zhang2025materialrefgs,ye20243d,yao2024reflective}.
Despite these advances, the above methods remain largely confined to static 3D modeling and lack the ability to capture and represent 4D dynamics.

\subsection{Dynamic Scene Modeling}

Modeling dynamic scenes from multi-view videos is a more challenging task that enables novel view synthesis across all time steps. The key difficulty lies in accurately representing the temporal motion of dynamic scenes. Inspired by the success of NeRF, a series of works~\cite{pumarola2021d,park2021nerfies,park2021hypernerf,li2022neural,fridovich2023k} extend Neural Radiance Fields to dynamic scene modeling by incorporating time awareness into the network. DyNeRF~\cite{li2022neural} introduces temporal conditioning into NeRF representations, enabling dynamic scene synthesis with radiance fields. To improve rendering efficiency, HexPlane~\cite{cao2023hexplane} and K-Planes~\cite{fridovich2023k} advance volumetric representations with compact feature planes, achieving faster dynamic NeRF training.

More recently, researchers have explored the potential of 3D Gaussian Splatting (3DGS) for 4D learning~\cite{luiten2024dynamic,li2024spacetime,zhu2024motiongs,shaw2024swings,liu2024modgs,bae2024per,lin2024gaussian,gao2024gaussianflow,lee2024fully,gao20257dgs,guo2024motion,liang2025gaufre,lu20243d,wang2025monofusion}. Dy3DGaussian~\cite{luiten2024dynamic} proposes a tracking-based approach that explicitly controls Gaussian motion over time. Works such as Deformable-3DGS~\cite{yang2024deformable} and 4DGaussians~\cite{wu20244d} adopt plane-feature–based architectures similar to dynamic NeRFs to model Gaussian deformation fields.
Different from these deformation-based approaches, 4D Gaussian Splatting (4DGS)~\cite{yang2023real} introduces a more direct dynamic representation by embedding time-aware attributes into Gaussian primitives. ST-GS~\cite{li2024spacetime} employs a hybrid scheme that combines temporally aware Gaussian representations with neural network-based appearance decoding. FreeTimeGS~\cite{wang2025freetimegs} further introduces motion attributes to enhance performance under complex dynamic motions. In addition, LongVolCap~\cite{xu2024representing} focuses on reconstructing long 4D video sequences, and Diffuman4D~\cite{jin2025diffuman4d} leverages diffusion-based priors for improved human motion modeling.

\section{Method}

\begin{figure*}
    \centering
    \includegraphics[width=\linewidth]{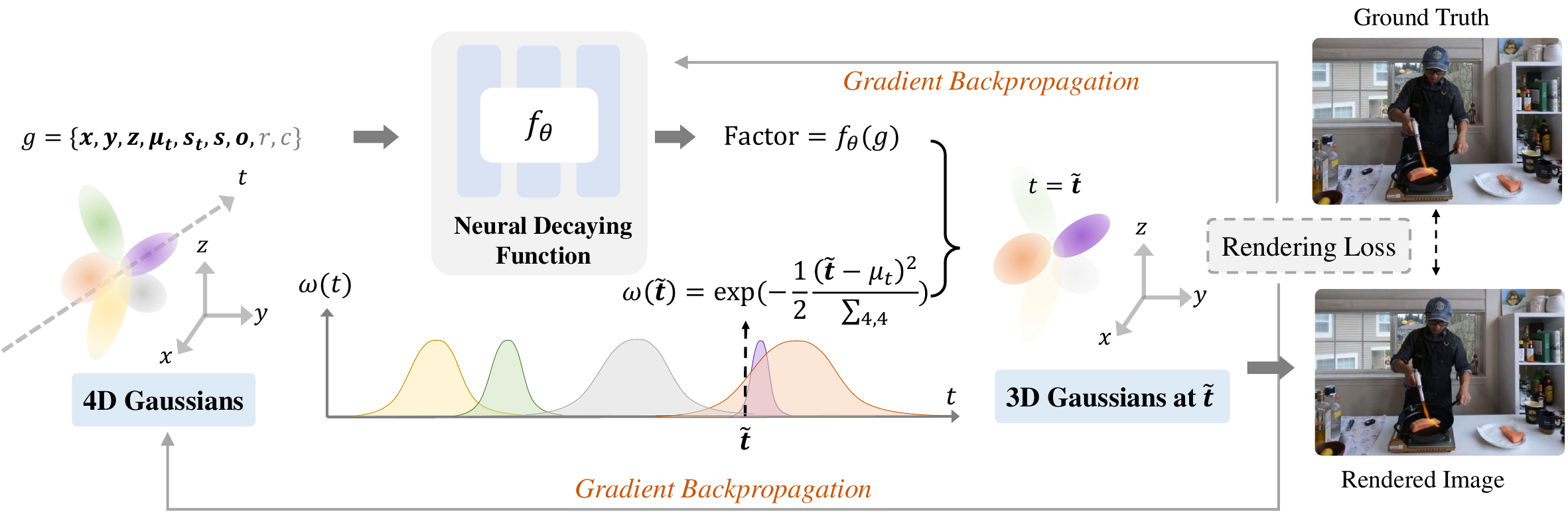}
    \vspace{-0.4cm}
    \caption{\textbf{Overview of 4C4D.}
    We introduce a Neural Decaying Function $f_\theta$, implemented as a lightweight neural network, to adaptively control the opacity decay of Gaussians. Given key Gaussian attributes as input, $f_\theta$ predicts a factor that controls the decay of Gaussian opacities. During training, both the Neural Decaying Function and the 4D Gaussians are jointly optimized via gradient backpropagation under a photometric rendering loss.}
    \label{fig:method}
    \vspace{-0.3cm}
\end{figure*}

We introduce 4C4D, a novel 4D Gaussian Splatting framework for reconstructing high-fidelity dynamic scenes from as few as four views. The overview of 4C4D is shown in Fig.~\ref{fig:method}. We begin with previewing 4D Gaussian Splatting in Sec.~\ref{sec.3.1}. Then, we introduce the methodology and implementation of Neural Decaying Function, as shown in Sec.~\ref{sec.3.2}. A hyper decay strategy that separately represents the decay function based on visibility detection is proposed in Sec.~\ref{sec.3.3}.

\subsection{Preliminary: 4D Gaussian Splatting}
\label{sec.3.1}
4D Gaussian Splatting (4DGS) represents a dynamic 4D scene using a set of Gaussians, each parameterized by attributes that model the time-varying geometry and appearance of the scene. A 4DGS $G=\{g_{i}\}_{i=1}^N$, consisting of $N$ 4D Gaussians, represents each Gaussian with key attributes: position $\sigma=\{x,y,z\}$, rotation $r$, opacity $o$, scale $s$, spherical harmonics coefficients $c$ and the temporal attributes $\mu_t$ and $s_t$, which define the timestep and duration of the 4D Gaussian, respectively. 

Similar to 3DGS, 4DGS formulates the 4D covariance matrix as $\sum = RSS^TR^T$, which can be factored in to a scale matrix $S\in \mathbb{R}$ and a rotation matrix $R \in \mathbb{R}^4$. The temporal attributes govern the evolution of the geometry and appearance of each 4D Gaussian at different timesteps.
Specifically, the position of the 4D Gaussian $g$ at timestep $\tilde{t}$ is given by:

\begin{equation}
    \sigma(\tilde{t}) = \sigma + {\sum\textstyle_{1:3,4}}{\sum\textstyle_{4,4}^{-1} (\tilde{t}-\mu_t)}.
\end{equation}

The opacity $o$ of 4D Gaussian $g$ at timestep $\tilde{t}$ can be formulated as:

\begin{equation}
        o(\tilde{t}) = \omega(\tilde{t}) * o,
\end{equation}
\begin{equation}
     \omega(\tilde{t}) = \text{exp}(-\frac{1}{2}\frac{(\tilde{t}-\mu_t)^2}{\sum\textstyle_{4,4}}),
\end{equation}
where $\omega(\tilde{t})$ represents a temporal opacity factor that controls the impact of each 4D Gaussian over time. In addition to the above geometric attributes, 4DGS models the appearances of each Gaussian using 4D spherical harmonics coefficients $c$. The coefficients model the color as $c(t,\theta,\phi)$, where $(\theta,\phi)$ represents the normalized view direction. The 4D spherical harmonics coefficients are formulated as:

\begin{equation}
    c(t,\theta,\phi) = \text{cos}(\frac{2\pi n}{T}t)Y_{lm}(\theta,\phi),
\end{equation}
where $Y_{lm}(\theta,\phi)$ is the spherical harmonics basis function with direction $(\theta,\phi)$, and $n$ donates the order of the Fourier series.

\begin{figure}
    \centering
    \includegraphics[width=\linewidth]{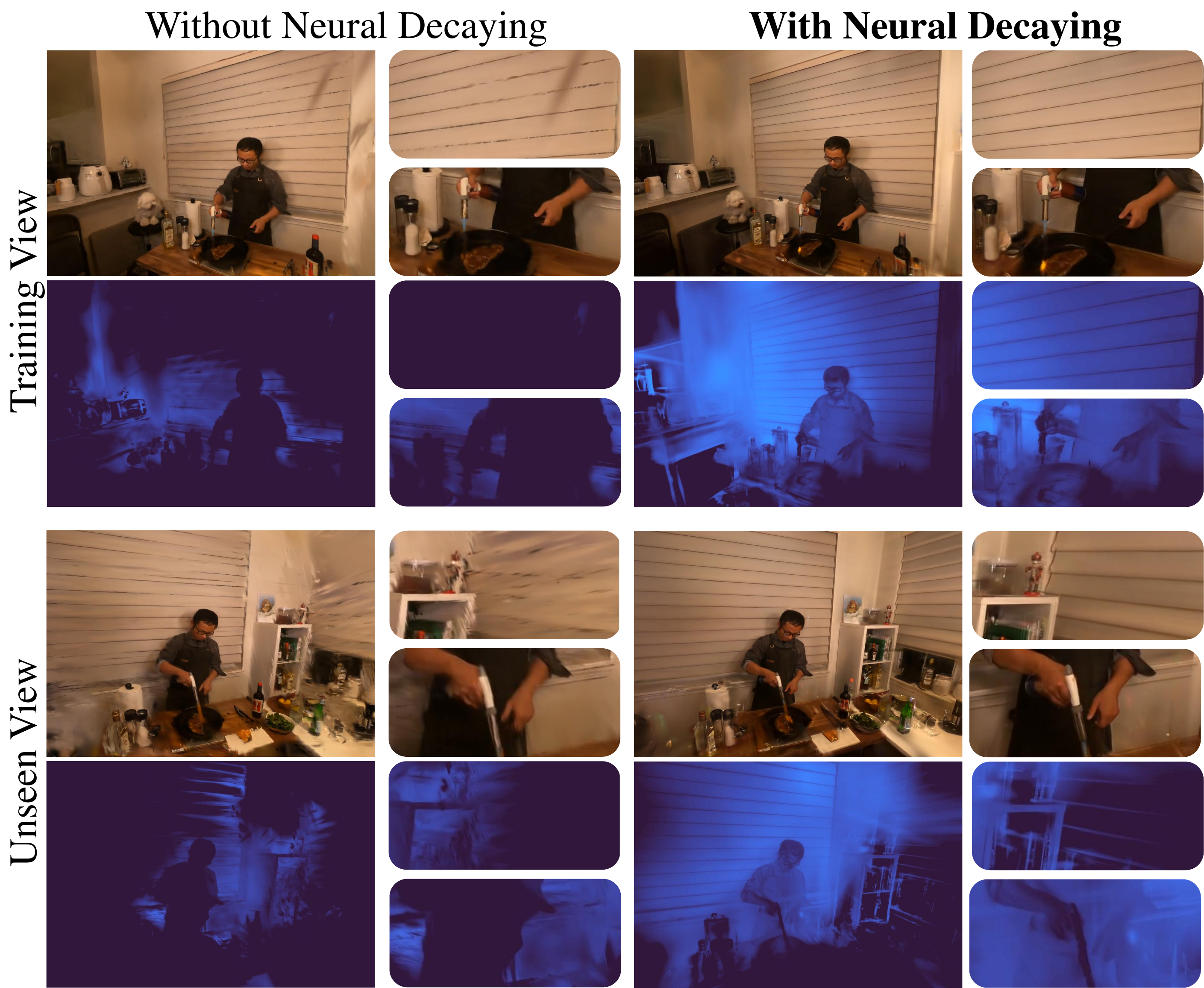}
    \caption{Rendered \textcolor{YellowOrange}{\textbf{RGB}} and \textcolor{RoyalBlue}{\textbf{depth}} results from 4DGS trained with and without our Neural Decaying Function, evaluated on both training and unseen views. Without neural decaying, 4DGS can overfit the training views but fails to generalize to unseen viewpoints, due to the learned poor geometry.}
    \vspace{-0.3cm}
    \label{fig:illustration}
\end{figure}

\subsection{Neural Decaying Function}
\label{sec.3.2}

We begin by analyzing the core challenge of 4D Gaussian Splatting (4DGS) under sparse-view conditions, which is the biased optimization that tends to overfit to appearance cues while neglecting geometric consistency. We show an example of training 4D Gaussian Splatting using only four cameras in Fig.~\ref{fig:illustration}. The figure shows the rendered geometry (depth) and appearance (RGB color) from the learned 4DGS under the training and testing views, respectively. As illustrated, 4DGS accurately reproduces the appearances at the training views but struggles to produce clean and consistent depth geometry due to insufficient spatial supervision, leading to noticeable artifacts and poor generalization to unseen viewpoints.

Driven by this observation, we introduce a neural opacity-decaying control that guides the optimization of 4DGS toward more faithful geometric learning. Specifically, we design a Neural Decaying Function $f_{\theta}$ implemented with a simple neural network. Given key attributes of each 4D Gaussian primitive, including its center $\sigma=\{x,y,z\}$, opacity $o$, and rotation $r$, the function predicts a decay factor $\tau$ for the Gaussian:

\begin{equation}
    \tau = f_\theta(x,y,z,o,r).
\end{equation}

The predicted $\tau$ neurally modulates Gaussian opacity, and the final opacity at timestep $\tilde{t}$ in 4C4D can be achieved with:

\begin{equation}
    o(\tilde{t}) = \tau * \text{exp}(-\frac{1}{2}\frac{(\tilde{t}-\mu_t)^2}{\sum\textstyle_{4,4}}) * o.
\end{equation}

This neural decaying mechanism redistributes the optimization gradients to emphasize geometric learning, thereby mitigating the overfitting bias and balancing geometry fidelity with appearance modeling. As shown in Figs.~\ref{fig:illustration}, 4C4D achieves both geometrically consistent depth reconstructions and visually appealing appearances across training and novel testing views.

\subsection{Separate Decay Strategy based on Visibility}
\label{sec.3.3}
We further justify that simply adopting the Neural Decaying Function to all 4D Gaussians lead to suboptimal training. The reason is that the gradients of 4DGS exist only in the visible regions that contribute to rendering at current view. Applying $f_\theta$ to invisible Gaussians distorts the optimization and results in performance degradation. To address this issue, we introduce a visibility detection strategy that operates in both spatial and temporal domains to identify visible 4D Gaussians at the current view and timestep. We then apply $f_\theta$ only to the visible regions while adopting a separate decay strategy for invisible 4D Gaussians. This ensures that the Gaussian gradients correctly update both the Neural Decaying Function and the 4D Gaussian primitives during optimization. The visibility detection of 4DGS $G$ at camera view $\tilde{v}$ and timestep $\tilde{t}$ is formulated as:

\begin{equation}
    G_{m} = Z_V(\tilde{v}, {\sigma}, Z_T(\tilde{t}, s_t, G)),
\end{equation}
where $Z_V$ and $Z_T$ donate the visibility detection based on view intersection and time intersection, respectively. $Z_V$ considers the 4D Gaussian centers $\sigma$ and filters out the Gaussians whose centers are not visible at current view. $Z_T$ considers the temporal span $s_t$ of each Gaussian and filters out those Gaussians whose duration does not include $\tilde{t}$.

\begin{figure*}[t]
    \centering
    \vspace{-0.1cm}
    \includegraphics[width=\linewidth]{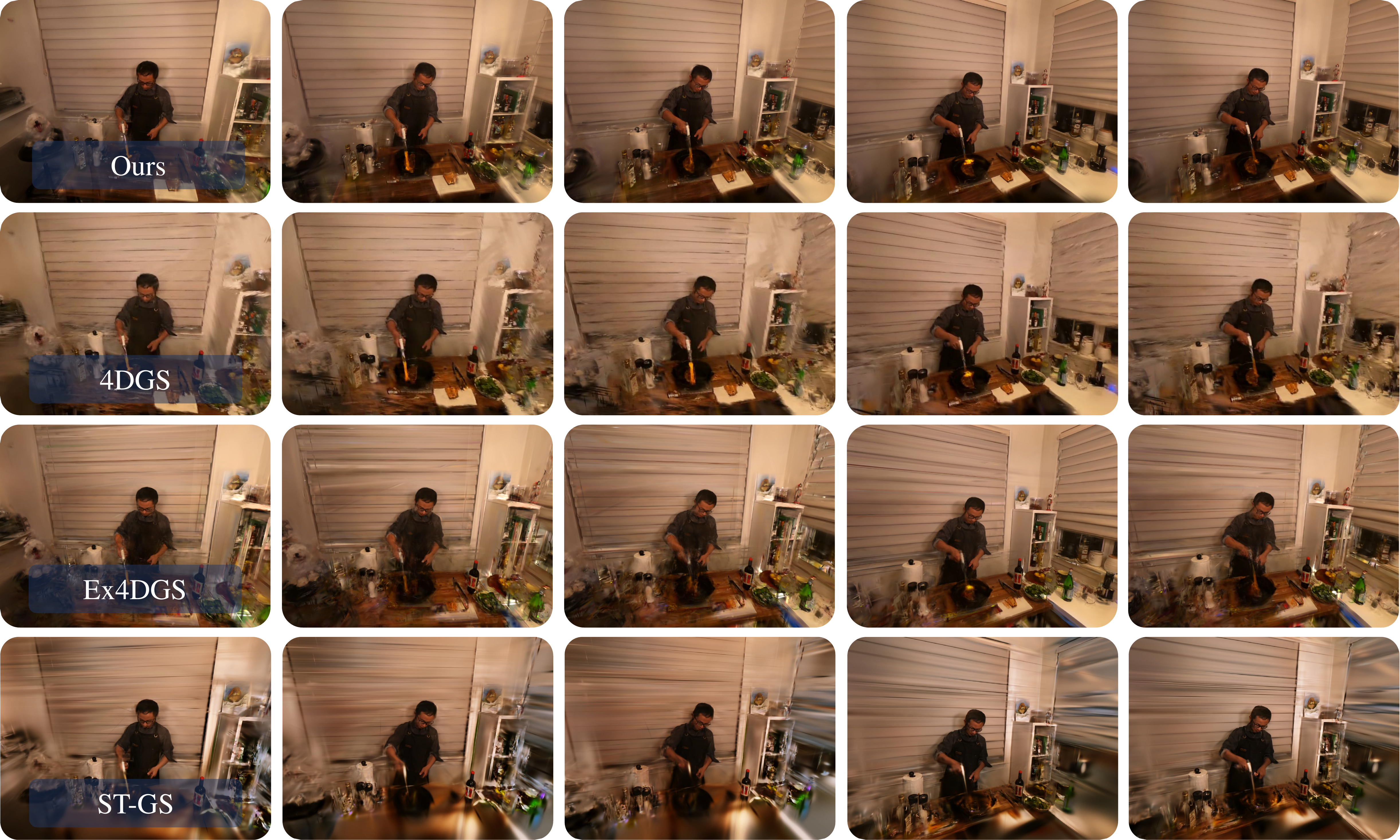}
    \caption{Visual comparisons under Neural3DV dataset.}
    \vspace{-0.4cm}
    \label{fig:n3v}
\end{figure*}

Through the visibility detection, we now obtain the visible 4D Gaussians $G_m$, while the invisible ones that do not contributes to Gaussian rendering at the current view and timestep are defined as $G_{{m}}^* = G - G_m$. We further find that applying a small constant decay to the opacity of invisible 4D Gaussians stabilizes training, consistent with the observations in \cite{han2024binocular}. The final separate decay strategy for each 4D Gaussian $g$ at the full set of 4DGS is defined as:

\begin{equation}
    \tau (g) =
        \begin{cases}
            f_\theta(x,y,z,o,r) & \text{if } g \in G_m, \\
            \beta & \text{if } g \in G_{{m}}^*,
        \end{cases}
\end{equation}
where the decay factor $\beta$ for invisible Gaussians is set to $0.999$.

\begin{figure*}
    \centering
    \includegraphics[width=\linewidth]{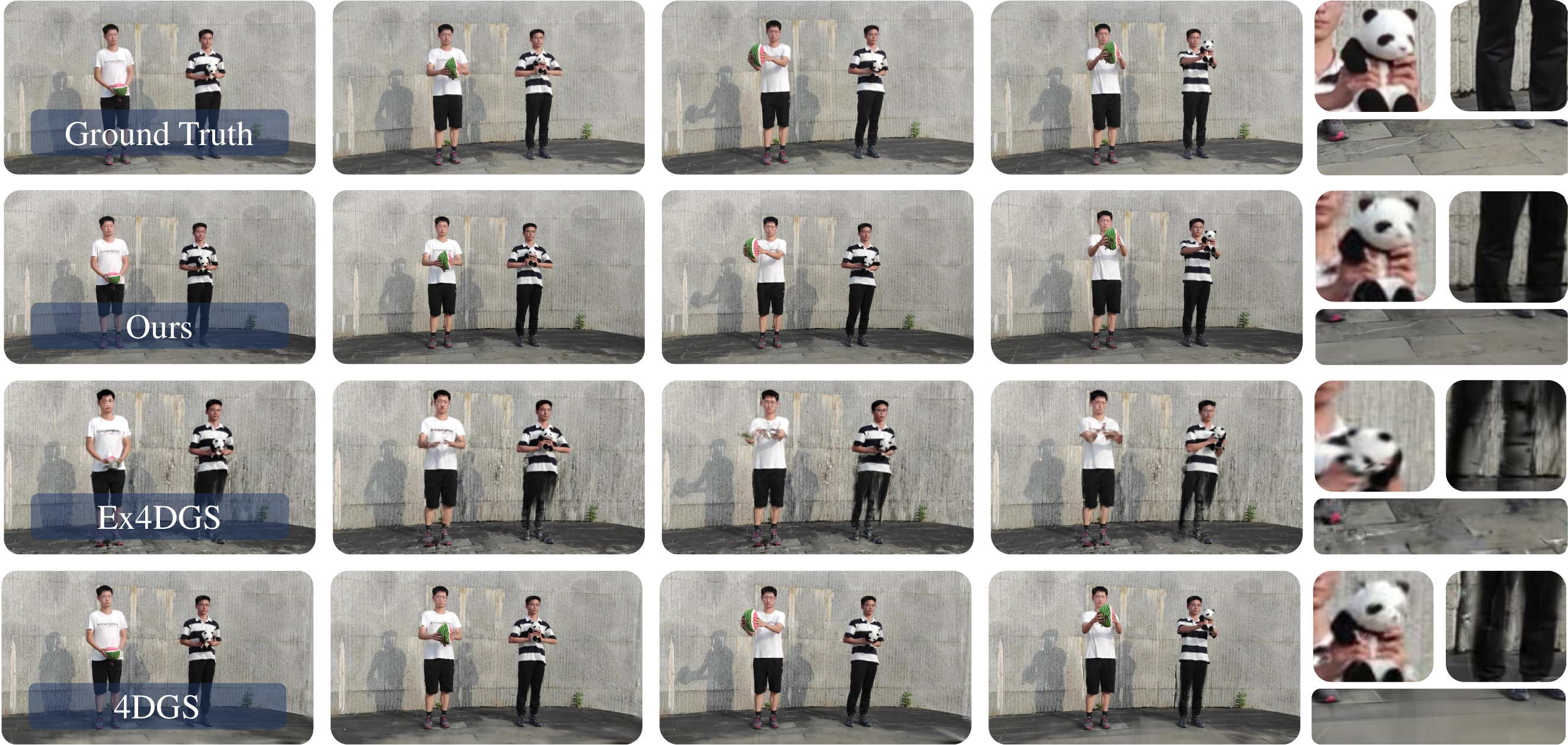}
    \vspace{-0.4cm}
    \caption{Visual comparisons under ENeRF-Outdoor dataset.}
    \vspace{-0.4cm}
    \label{fig:enerf}
\end{figure*}

\section{Experiment}

\subsection{Setup}

\noindent\textbf{Dataset.}
We conduct comprehensive experiments on four datasets to evaluate 4C4D, including Neural3DV~\cite{li2022neural}, ENeRF-Outdoor~\cite{lin2022efficient}, Mobile-Stage~\cite{xu20244k4d}, and our self-captured Dyn4Cam dataset.
Neural3DV~\cite{li2022neural} is a widely used multi-view video dataset containing six dynamic scenes. Each scene is recorded with 18–21 cameras at a resolution of 2704×2028 and 30 FPS. Following prior works, we downsample the images by a factor of 0.5 before training, and we use the first 300 frames of each scene for both training and evaluation. For the sparse-view setting, we select four cameras located at the most distant positions as the training views, while the remaining camera videos are used as test views.
ENeRF-Outdoor~\cite{lin2022efficient} contains three outdoor dynamic scenes captured by an 18-camera array. The videos are recorded at a resolution of 1920×1080 and 60 FPS. We adopt the same four-camera train/test split strategy as in Neural3DV.
Mobile-Stage, released by 4K4D~\cite{xu20244k4d}, contains a complex scene featuring highly dynamic multi-person dance performances. The scene is captured using 24 cameras at a resolution of 1920×1080. We select four views with large overlap for training, and use the remaining views as test data.

Finally, since all existing datasets are densely captured using camera arrays with around 20 devices, we additionally collect Dyn4Cam, a new dataset captured with only four portable GoPro-style cameras. The videos are recorded at 60 FPS with a resolution of 1920×1080, and each scene contains 300 frames. As Dyn4Cam does not include held-out test views, we only provide qualitative comparisons on this dataset.

\noindent\textbf{Metrics.} Following prior works~\cite{yang2023real,li2024spacetime,wang2025freetimegs}, we report PSNR, DSSIM and LPIPS as our evaluation metrics for assessing the visual quality of novel view rendering. PSNR is the most commonly used pixel-wise metric that measures the per-pixel $\ell_2$ difference between the rendered image and the ground-truth image. DSSIM evaluates the structural dissimilarity between two images, and we report two variants, DSSIM$_1$ and DSSIM$_2$, computed with different data-range settings of 1.0 and 2.0. LPIPS measures perceptual similarity between images, and lower values indicate better perceptual quality.

\noindent\textbf{Baselines.} We compare 4C4D with state-of-the-art Gaussian-based baselines including 4DGaussians~\cite{wu20244d}, Ex4DGS~\cite{lee2024fully}, ST-GS~\cite{li2024spacetime}, and 4DGS~\cite{yang2023real}. All baselines follow the same experimental setup as 4C4D to ensure a fair comparison.

\begin{table}[t]
\centering
\caption{Quantitative comparison under Neural3DV dataset.}
\resizebox{\linewidth}{!}{
\begin{tabular}{lcccc}
\toprule
Method & PSNR $\uparrow$ & DSSIM$_1$ $\downarrow$& DSSIM$_2$ $\downarrow$ & LPIPS $\downarrow$ \\
\midrule
4DGaussians & \cellcolor{orange}20.82 & \cellcolor{orange}0.117 & \cellcolor{orange}0.077 & \cellcolor{orange}0.190 \\
STGS            & 17.70 & 0.158  & 0.107  & 0.325  \\
4DGS            & 20.60 & 0.143 & 0.094 & 0.244 \\
Ex4DGS          & 19.33 & 0.149 & 0.099 & 0.239 \\
\midrule
Ours            & \cellcolor{red}22.29 & \cellcolor{red}0.098 & \cellcolor{red}0.062 & \cellcolor{red}0.146 \\
\bottomrule
\end{tabular}}
\vspace{-0.4cm}
\label{tab:n3v}
\end{table}

\begin{figure*}
    \centering
    \includegraphics[width=\linewidth]{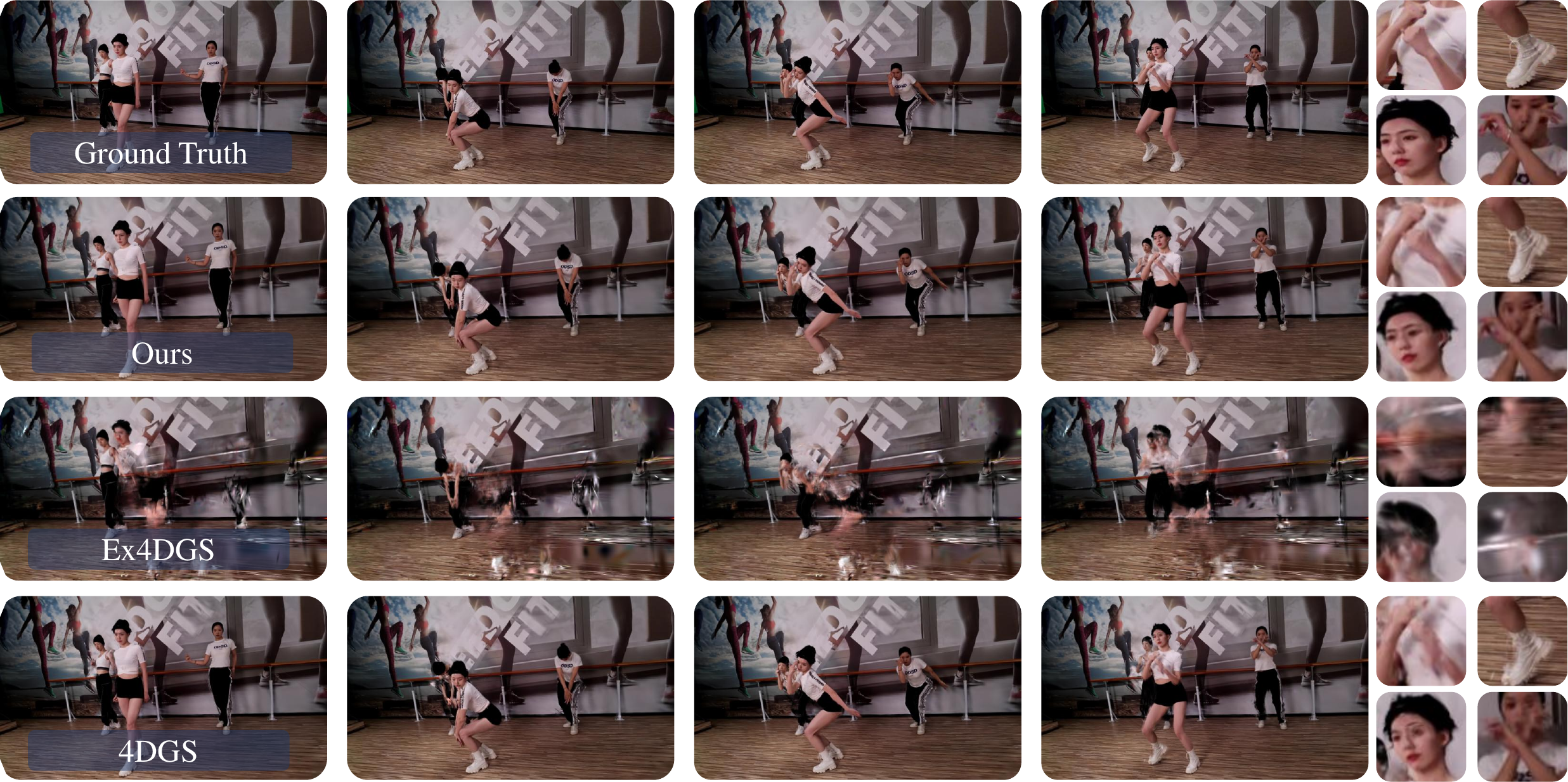}
    \vspace{-0.2cm}
    \caption{Visual comparisons under Mobile-Stage dataset.}
    \label{fig:mobile}
\vspace{-0.4cm}
\end{figure*}

\subsection{Results of 4-Camera Novel View Synthesis}
\noindent\textbf{Neural3DV.} We conduct comprehensive evaluations on the Neural3DV dataset. As shown in Tab.~\ref{tab:n3v}, 4C4D consistently outperforms all state-of-the-art baselines across all quantitative metrics, demonstrating clear advantages in both geometric accuracy and photometric quality. Visual comparisons in Fig.~\ref{fig:n3v} further highlight the superiority of our reconstructions. Under the challenging sparse-view setting with supervision from only four cameras, most baselines exhibit unstable 4D reconstruction behaviors, including noisy point distributions, drifting primitives, and noticeable Gaussian artifacts.

In contrast, 4C4D produces clean and temporally coherent 4D Gaussian structures, yielding high-fidelity appearance and robust geometry even in regions with limited viewpoint coverage. This enables visually compelling novel view synthesis across a wide range of timesteps and viewpoints. The visualizations in Fig.~\ref{fig:n3v} follow a circular camera trajectory over time, showing the reconstructed scenes at multiple timestamps and perspectives, and clearly revealing improvements in temporal stability, geometric completeness, and rendering consistency achieved by our approach.

\begin{table}[t]
\centering
\caption{Quantitative comparisons under ENeRF-Outdoor dataset.}
\resizebox{\linewidth}{!}{
\begin{tabular}{lcccc}
\toprule
Method & PSNR $\uparrow$ & DSSIM$_1$ $\downarrow$ & DSSIM$_2$ $\downarrow$ & LPIPS $\downarrow$ \\
\midrule
4DGaussians & 18.21 & 0.300    & 0.183    & 0.456 \\
STGS            & 15.51 & 0.315 & 0.186 & 0.688 \\
4DGS            & \cellcolor{orange}23.52 & \cellcolor{orange}0.183 & \cellcolor{orange}0.112 & \cellcolor{orange}0.151 \\
Ex4DGS          & 21.89 & 0.224 & 0.138 & 0.263 \\
\midrule
Ours            & \cellcolor{red}24.32 & \cellcolor{red}0.162 & \cellcolor{red}0.097 & \cellcolor{red}0.121 \\
\bottomrule
\end{tabular}}
\vspace{-0.4cm}
\label{tab:enerf}
\end{table}

\noindent\textbf{ENeRF-Outdoor.} We further evaluate 4C4D on the ENeRF-Outdoor dataset, which contains real-world scenes characterized by complex object motions, uncontrolled lighting conditions, and significant background clutter. These outdoor environments introduce strong illumination variations, shadows, and reflections, making dynamic 4D reconstruction especially challenging. Quantitative results in Tab.~\ref{tab:enerf} demonstrate that 4C4D substantially outperforms all baseline methods, achieving consistently superior performance across all evaluation metrics. 

Detailed visualizations in Fig.~\ref{fig:enerf} reveal that 4C4D reconstructs cleaner and more detailed 4D scenes, resulting in noticeably higher rendering quality. In particular, 4C4D exhibits strong robustness in modeling rapid motions from sparse-view videos, successfully capturing the dynamics of objects such as the doll and the hands. Moreover, 4C4D consistently improves the rendering quality at the static regions like the grounds.

\begin{table}[t]
\caption{Quantitative comparison under Mobile-Stage dataset.}
\vspace{-0.2cm}
\centering
\resizebox{\linewidth}{!}{
\begin{tabular}{lcccc}
\toprule
Method & PSNR $\uparrow$ & DSSIM$_1$ $\downarrow$ & DSSIM$_2$ $\downarrow$ & LPIPS $\downarrow$ \\
\midrule
4DGaussians & 20.15 & 0.218 & 0.144 & 0.226 \\
STGS            & 14.30 & 0.321 & 0.229 & 0.658 \\
4DGS            & \cellcolor{orange}22.15 & \cellcolor{orange}0.199 & \cellcolor{orange}0.135 & \cellcolor{orange}0.180 \\
Ex4DGS & 17.85 & 0.272 & 0.193 & 0.260 \\
\midrule
Ours            & \cellcolor{red}22.36 & \cellcolor{red}0.194 & \cellcolor{red}0.129 & \cellcolor{red}0.121 \\
\bottomrule
\end{tabular}}
\vspace{-0.4cm}
\label{tab:mobile}
\end{table}

\noindent\textbf{Mobile-Stage.} 
We further evaluate 4C4D on the Mobile-Stage dataset, which contains large-scale outdoor scenes with multiple actors performing highly dynamic and coordinated dance motions. This dataset presents substantial challenges for 4D reconstruction, including rapid limb movements, complex multi-person interactions, frequent self-occlusions, and cluttered backgrounds. Qualitative and quantitative comparisons are provided in Fig.~\ref{fig:mobile} and Tab.~\ref{tab:mobile}.

As shown in Tab.~\ref{tab:mobile}, our method achieves the strongest performance across all evaluation metrics. The visualizations in Fig.~\ref{fig:mobile} further highlight that 4C4D captures both high-fidelity geometry and consistent appearance throughout the entire dynamic sequence. Importantly, while all competing baselines struggle to reconstruct the subtle and fast motion details in the three-person dance performance, 4C4D successfully recovers these fine-scale dynamics from only four input cameras. This result demonstrates the robustness of 4C4D in challenging real-world multi-person scenarios and its ability to generalize effectively to large-scale dynamic scenes.

\begin{figure*}
    \centering
    \includegraphics[width=\linewidth]{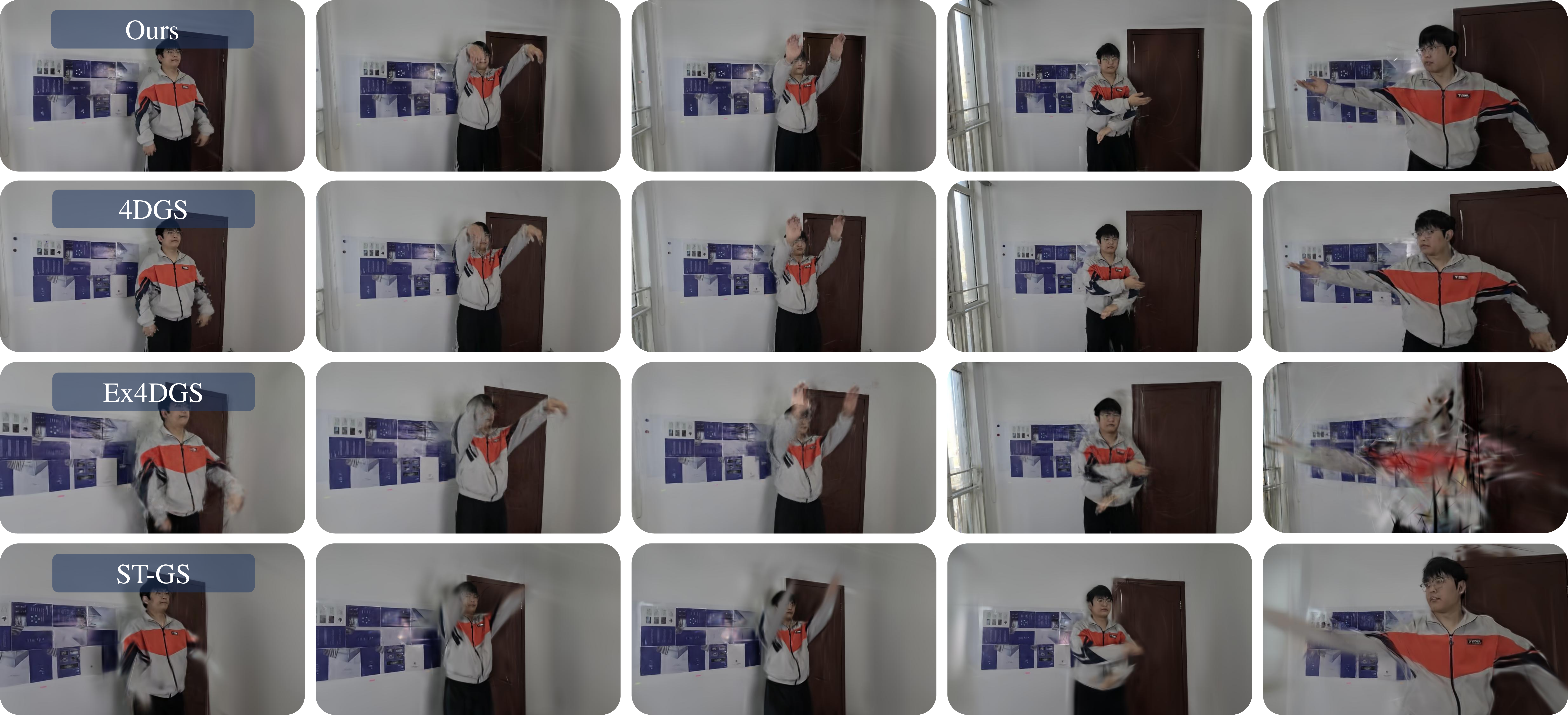}
\vspace{-0.5cm}
    \caption{Visual comparisons under the self-captured Dyn4Cam dataset.}
    \label{fig:selfcap}
    \vspace{-0.4cm}
\end{figure*}

\noindent\textbf{Dyn4Cam.} To further demonstrate the effectiveness of 4C4D on multi-view videos captured with sparse and fully portable cameras, we additionally evaluate our method on Dyn4Cam, a self-captured dynamic scene dataset. Dyn4Cam is recorded using only four GoPro-style consumer cameras that are inexpensive and easy for everyday users to deploy, with the entire capture setup costing less than 1,500 USD.

As shown in Fig.~\ref{fig:selfcap}, 4C4D generates more temporally consistent 4D dynamics and produces significantly more faithful and stable dynamic novel-view renderings compared to previous methods. Because Dyn4Cam is collected using only four cameras and does not contain any extra viewpoints for numerical evaluation, we do not report quantitative metrics on this dataset. The visualizations in Fig.~\ref{fig:selfcap} follow a circular camera trajectory over time, illustrating the reconstructed scenes at various timesteps and viewpoints and highlighting the differences in temporal stability and geometric fidelity across methods.

\begin{table}[t]
\caption{Ablation study on the framework designs.}
\centering
\resizebox{\linewidth}{!}{
\begin{tabular}{lcccc}
\toprule
Method & PSNR $\uparrow$ & DSSIM$_1$ $\downarrow$ & DSSIM$_2$ $\downarrow$ & LPIPS $\downarrow$ \\
\midrule
w/o Neural Decay               & 22.60 & 0.097 & 0.062 & 0.147 \\
w/o Vis Detect           & \cellcolor{orange}24.49 & \cellcolor{orange}0.075 & \cellcolor{orange}0.045 & \cellcolor{orange}0.127 \\
Full & \cellcolor{red}24.68 & \cellcolor{red}0.070 & \cellcolor{red}0.042 & \cellcolor{red}0.115 \\
\bottomrule
\end{tabular}}
\label{tab:ablation1}
\end{table}

\begin{table}[t]
\caption{Ablation study on decaying functions.}
\centering
\resizebox{\linewidth}{!}{
\begin{tabular}{lcccc}
\toprule
Method & PSNR $\uparrow$ & DSSIM$_1$ $\downarrow$ & DSSIM$_2$ $\downarrow$ & LPIPS $\downarrow$ \\
\midrule
None               & 22.60 & 0.097 & 0.062 & 0.147 \\
Constant         & 24.31 & 0.075 & 0.046 & 0.125 \\
Pow Decay           & \cellcolor{orange}24.35 & \cellcolor{orange}0.074 & \cellcolor{orange}0.045 & \cellcolor{orange}0.124 \\
Exp Decay           & 24.32 & 0.077 & 0.047 & 0.135 \\
Ours & \cellcolor{red}24.68 & \cellcolor{red}0.070 & \cellcolor{red}0.042 & \cellcolor{red}0.115 \\
\bottomrule
\end{tabular}}
\label{tab:ablation2}
\end{table}

\subsection{Ablation Study}

\noindent\textbf{Framework Designs.}
To evaluate the effectiveness of the major components in our method, we conduct a series of ablation studies on the representative data in Neural3DV dataset and report the quantitative performance in Tab.~\ref{tab:ablation1}. We first assess the contribution of the proposed Neural Decaying Function by removing the learnable decay module and replacing it with a fixed decay scheme, reported as ``w/o Neural Decay''. The result demonstrates that an optimal opacity decay that helps the geometric learning in 4DGS plays a crucial in improving dynamic modeling capabilities. We then examine the importance of the visibility detection strategy by disabling the visibility-aware masking and applying the same decay function uniformly to all regions, reported as “w/o Vis Detect”.  The ablation results consistently validate the effectiveness of our design choices. Both components contribute notably to improving geometric stability and photometric consistency, and their combined effect leads to substantial gains in dynamic novel view synthesis quality.

\noindent\textbf{Variant Decaying Functions.}
To further demonstrate that our proposed Neural Decaying Function, which is learned during training, provides an optimal opacity decay scheme, we conduct additional ablation studies by replacing it with several alternative designs. The results are shown in Tab.~\ref{tab:ablation2}. The most naive baseline is a constant opacity decay, as used in \cite{han2024binocular}. We adopt the same decay coefficient of 0.9 and report the result as ``Constant''. We also explore several manually designed mathematical functions that compute the decay coefficient based on the current opacity. For example, (1) we define a hand-tuned exponential function and report the result as ``Exp Decay'', and (2) we define a hand-tuned power function and report the result as ``Pow Decay''. The ablation results show that the learnable Neural Decaying Function consistently outperforms all manually crafted alternatives, even those requiring extensive effort to tune, demonstrating its ability to automatically learn an optimal opacity decay strategy.

\section{Conclusion}

We introduce 4C4D, a novel framework that achieves high-fidelity 4D Gaussian Splatting from videos captured by as few as four portable cameras. The core idea of 4C4D is a Neural Decaying Function that improves geometric modeling by learning to decay Gaussian opacities adaptively. By directing the 4DGS gradients to focus more on geometry, this mechanism alleviates the inherent imbalance between geometry and appearance learning in 4DGS and results in less biased optimization. We conduct comprehensive evaluations on diverse sparse-view dynamic scene datasets with varying camera overlaps and demonstrate that 4C4D consistently outperforms state-of-the-art methods.

\section{Acknowledgment}

This work was partially supported by Deep Earth Probe and Mineral Resources Exploration -- National Science and Technology Major Project (2024ZD1003405), and the National Natural Science Foundation of China (62272263, 625B2102), and in part by Kuaishou. Junsheng Zhou is also partially funded by China Association for Science and Technology Young Talents Support Project (Doctoral Program) and Baidu Scholarship.

{
    \small
    \bibliographystyle{ieeenat_fullname}
    \bibliography{main}
}

\setcounter{page}{1}
\maketitlesupplementary
\renewcommand\thesection{\Alph{section}}
\setcounter{section}{0}

\section{Details of Capture System}

\subsection{Hardware}

We build our capture system using four GoPro HERO 12 cameras, a modern consumer-grade portable device that is easily accessible to ordinary users. The entire setup, including the cameras and custom mounting brackets, costs less than 1,500 USD. All four cameras record video with identical settings at a resolution of 1920$\times$1080 and a frame rate of 60 frames per second.

\subsection{Capturing Pipeline}

We position four cameras facing the region where the dynamic actions occur. The four viewpoints span an effective coverage of approximately 100–120 degrees, providing sufficient overlap to fully observe the scene without causing the severe difficulties associated with non-overlapping views. Due to the inherent temporal misalignment of GoPro devices, we first perform multi-view frame re-alignment to ensure perfect synchronization across all videos.

To obtain accurate camera poses, we use COLMAP~\cite{schonberger2016structure} to solve the extrinsics before recording videos. Estimating stable poses from only four photos can often be unreliable. Therefore, we capture an additional set of eight images of the same scene to improve multi-view feature matching and enhance pose stability. This step does not involve any extra cameras, where the same four cameras are used to take photos from additional viewpoints. After computing all camera poses, only the four main viewpoints are used for video capture, and only their corresponding poses are utilized in the subsequent 4D Gaussian Splatting training.

\subsection{The Dyn4Cam Dataset}

We finally collect eight representative action categories to form the Dyn4Cam dataset, including Boxing, Dancing, Chest Fly, Exercising, Jumping Clap, Running, Taichi, and Showing a Toy. These actions span both fast and relatively slow motions and include simple daily movements as well as complex athletic behaviors. This composition ensures good diversity and provides a comprehensive benchmark for dynamic scene reconstruction in sparse views.

\section{Implementation Details}
 
We implement 4C4D in PyTorch and adopt the Adam optimizer~\cite{kingma2014adam} to optimize the Gaussian primitives. We adopt the MAST3R~\cite{leroy2024grounding} for initialization, which is more stable in the sparse-camera setting. To ensure stable warm-up, we introduce the Neural Decaying Function after 500 iterations, which provides more reliable supervision to the neural network and helps stabilize the optimization process. 

\section{More Experimental Results}

\subsection{Results on Neural3DV dataset}

\begin{figure*}
    \centering
    \includegraphics[width=0.92\linewidth]{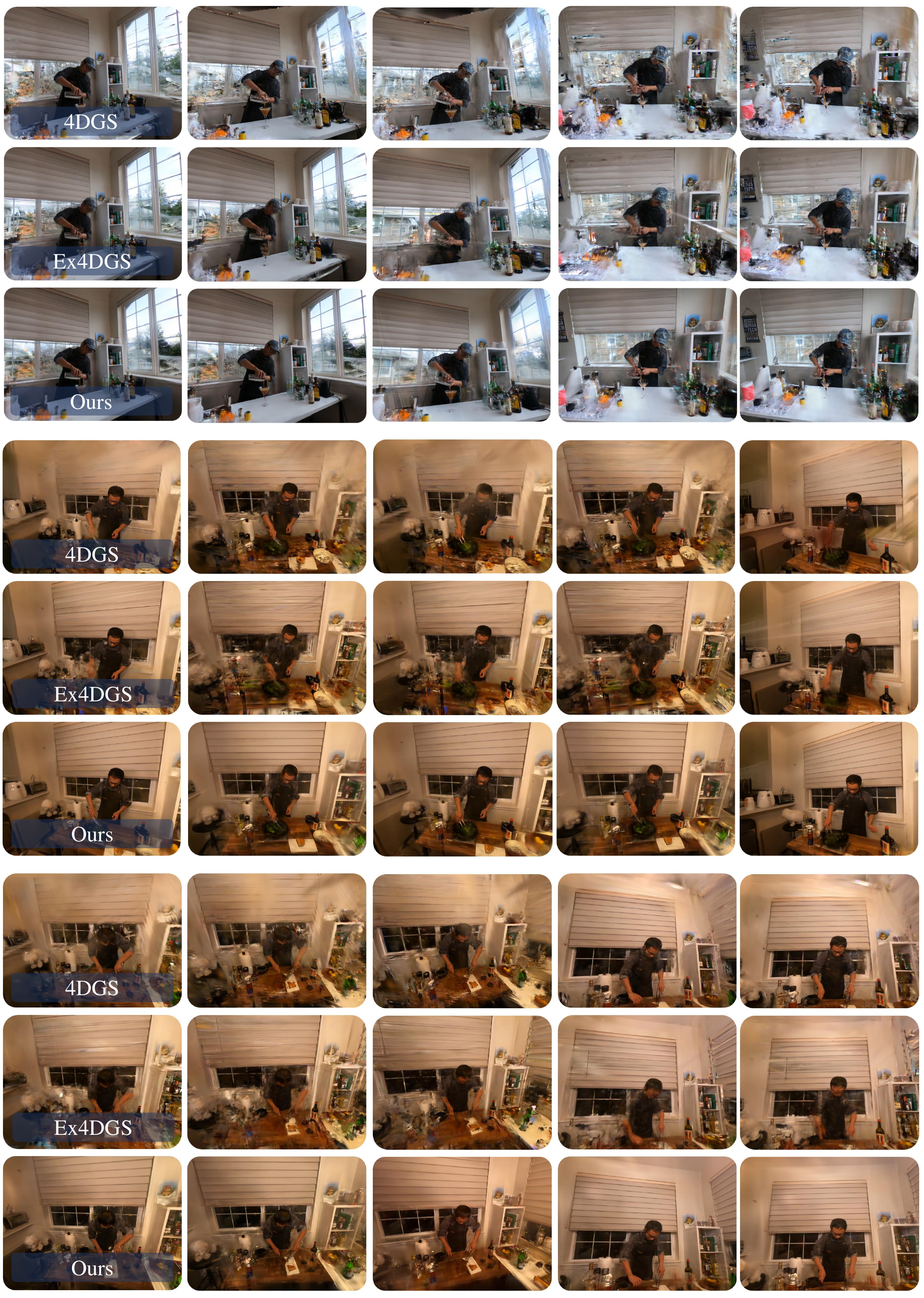}
    \caption{More visual comparisons under Neural3DV dataset.}
    \label{fig:n3v_supp}
\end{figure*}

We report the complete evaluation metrics for each scene in the Neural3DV dataset in Tab.~\ref{tab:group1} and Tab.~\ref{tab:group2}. 
Since the 'Flame Salmon' scene is significantly longer than the other five scenes, we restrict the evaluation to its first 300 frames to ensure a consistent experimental setup.
Additional visual comparisons are provided in Fig.~\ref{fig:n3v_supp}, where all baselines are rendered using the same camera trajectory.

\subsection{Results on ENeRF-Outdoor dataset}

\begin{figure*}
    \centering
    \includegraphics[width=1\linewidth]{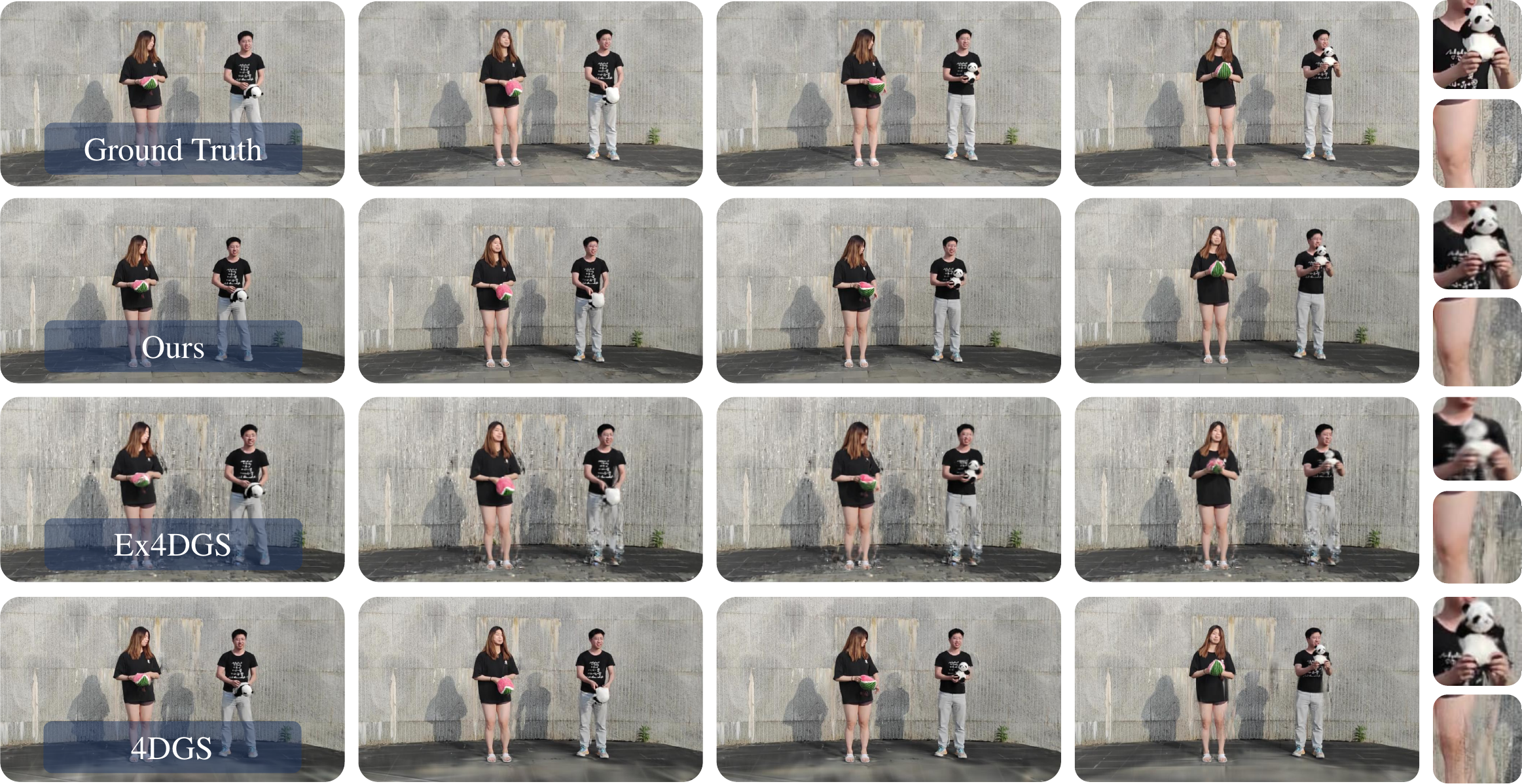}
    \caption{More visual comparisons under ENeRF-Outdoor dataset.}
    \label{fig:enerf_supp}
\end{figure*}

Tab.~\ref{tab:group3} presents a detailed breakdown of the per-scene quantitative results on the ENeRF-Outdoor dataset, showcasing the performance of all competing methods across a diverse set of outdoor sequences. To complement these numerical comparisons, additional qualitative visualizations are provided in Fig.~\ref{fig:enerf_supp}.

\subsection{Results on Dyn4Cam dataset}

\begin{figure*}
    \centering
    \includegraphics[width=\linewidth]{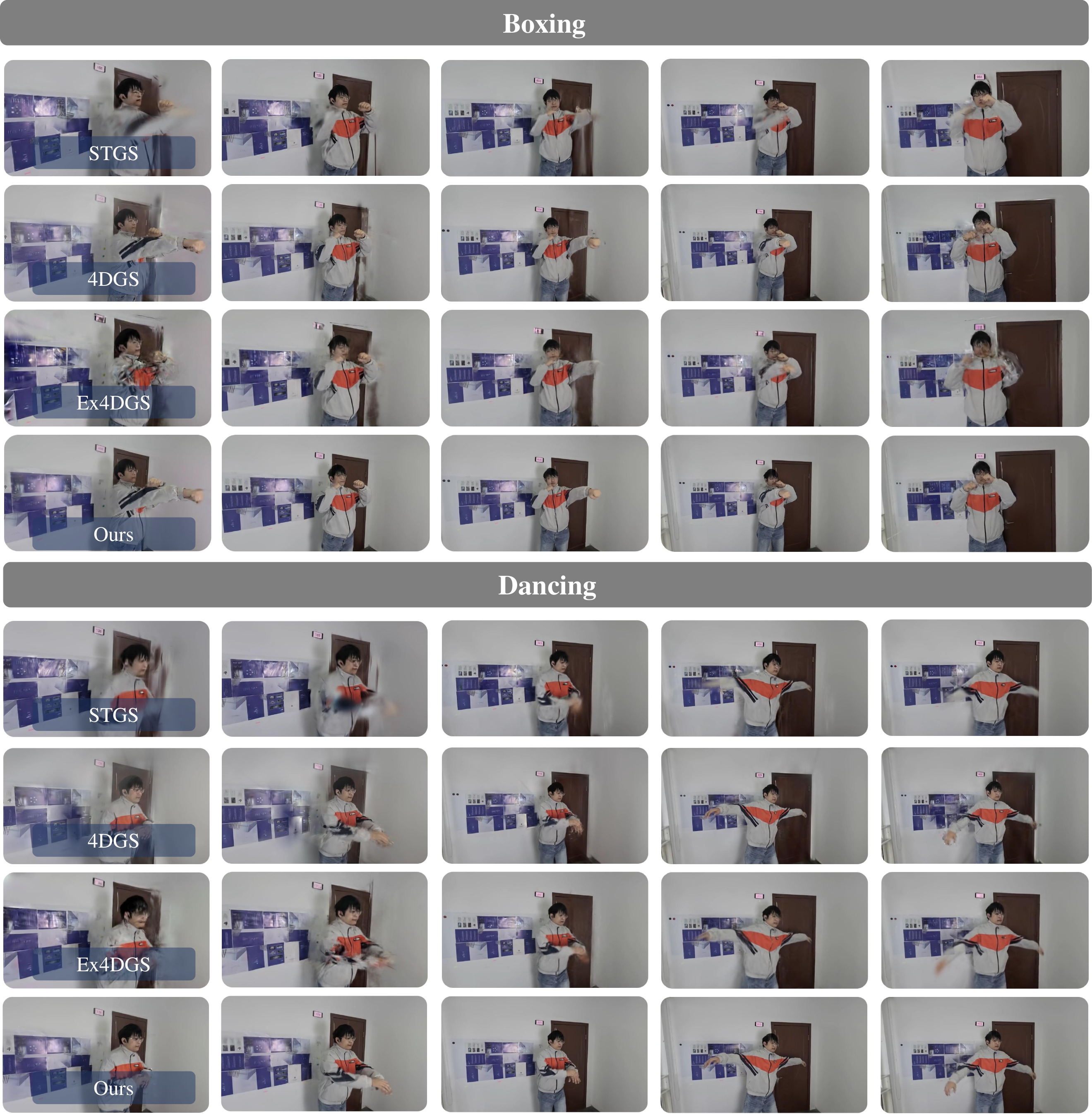}
    \caption{More visual comparisons under Dyn4Cam dataset (Part 1).}
    \label{fig:dyn4cam_1}
\end{figure*}

\begin{figure*}
    \centering
    \includegraphics[width=\linewidth]{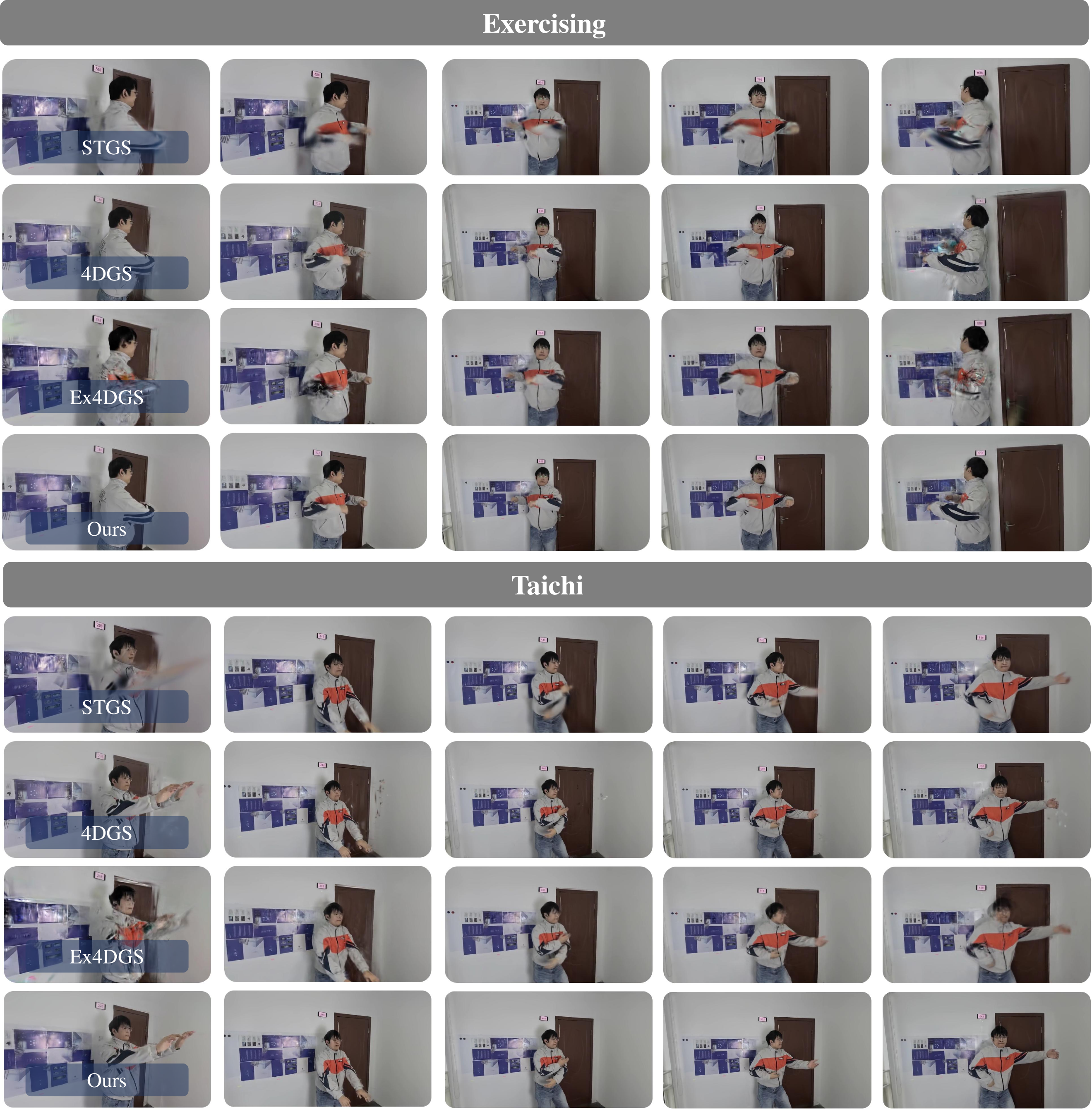}
    \caption{More visual comparisons under Dyn4Cam dataset (Part 2).}
    \label{fig:dyn4cam_2}
\end{figure*}

Additionally, we provide more comprehensive qualitative comparisons with state-of-the-art baselines in Fig.~\ref{fig:dyn4cam_1} and Fig.~\ref{fig:dyn4cam_2}. These visual examples cover four representative scenes from the Dyn4Cam dataset, namely “Taichi,” “Dancing,” “Boxing,” and “Exercising.” These sequences span a wide range of motion patterns and appearance variations. As shown in the figures, our approach consistently produces sharper details, more stable temporal behaviors, and cleaner geometry across different motion types.

\begin{table*}[t]
\centering
\resizebox{\linewidth}{!}{
\begin{tabular}{lcccccccccccc}
\toprule
& \multicolumn{4}{c}{Coffee Martini} 
& \multicolumn{4}{c}{Cook Spinach} 
& \multicolumn{4}{c}{Cut Roasted Beef} \\
\cmidrule(lr){2-5} \cmidrule(lr){6-9} \cmidrule(lr){10-13}
Method & PSNR & DSSIM$_1$ & DSSIM$_2$ & LPIPS 
& PSNR & DSSIM$_1$ & DSSIM$_2$ & LPIPS
& PSNR & DSSIM$_1$ & DSSIM$_2$ & LPIPS \\
\midrule
STGS~\cite{li2024spacetime}   & 16.62 & 0.189 & 0.129 & 0.373
       & 18.72 & 0.134 & 0.090 & 0.278
       & 18.65 & 0.135 & 0.089 & 0.286 \\
4DGS~\cite{yang2023real}   & \cellcolor{orange}17.50 & \cellcolor{orange}0.174 & \cellcolor{orange}0.118 & 0.289
       & \cellcolor{orange}21.96 & \cellcolor{orange}0.106 & \cellcolor{orange}0.067 & \cellcolor{orange}0.205
       & \cellcolor{orange}21.90 & \cellcolor{orange}0.102 & \cellcolor{orange}0.064 & \cellcolor{orange}0.191 \\
Ex4DGS~\cite{lee2024fully} & 17.45 & 0.179 & 0.124 & \cellcolor{orange}0.281
       & 20.61 & 0.138 & 0.089 & 0.224
       & 20.33 & 0.144 & 0.092 & 0.228 \\
\midrule
Ours   & \cellcolor{red}19.57 & \cellcolor{red}0.140 & \cellcolor{red}0.093 & \cellcolor{red}0.203    & \cellcolor{red}24.44 & \cellcolor{red}0.075 & \cellcolor{red}0.046 & \cellcolor{red}0.111
       & \cellcolor{red}23.22 & \cellcolor{red}0.083 & \cellcolor{red}0.051 & \cellcolor{red}0.124 \\

\bottomrule
\end{tabular}}
\caption{Quantitative comparison on Neural3DV dataset (Part 1).}
\label{tab:group1}
\end{table*}

\begin{table*}[t]
\centering
\resizebox{\linewidth}{!}{
\begin{tabular}{lcccccccccccc}
\toprule
& \multicolumn{4}{c}{Flame Salmon} 
& \multicolumn{4}{c}{Flame Steak} 
& \multicolumn{4}{c}{Sear Steak} \\
\cmidrule(lr){2-5} \cmidrule(lr){6-9} \cmidrule(lr){10-13}
Method & PSNR & DSSIM$_1$ & DSSIM$_2$ & LPIPS 
& PSNR & DSSIM$_1$ & DSSIM$_2$ & LPIPS
& PSNR & DSSIM$_1$ & DSSIM$_2$ & LPIPS \\
\midrule
STGS~\cite{li2024spacetime}   & 14.54 & 0.227 & 0.162 & 0.474
       & 18.96 & 0.131 & 0.086 & 0.270
       & 18.72 & 0.136 & 0.090 & 0.271 \\
4DGS~\cite{yang2023real}   & \cellcolor{orange}17.81 & \cellcolor{orange}0.163 & \cellcolor{orange}0.110 & \cellcolor{orange}0.283
       & \cellcolor{orange}22.55 & 0.212 & 0.146 & 0.294
       & \cellcolor{orange}21.86 & \cellcolor{orange}0.102 & \cellcolor{orange}0.064 & 0.202 \\
Ex4DGS~\cite{lee2024fully} & 15.40 & 0.193 & 0.135 & 0.308
       & 21.39 & \cellcolor{orange}0.123 & \cellcolor{orange}0.077 & \cellcolor{orange}0.200
       & 20.80 & 0.119 & 0.078 & \cellcolor{orange}0.196 \\
\midrule
Ours   & \cellcolor{red}19.15 & \cellcolor{red}0.136 & \cellcolor{red}0.091 & \cellcolor{red}0.205
       & \cellcolor{red}23.72 & \cellcolor{red}0.075 & \cellcolor{red}0.046 & \cellcolor{red}0.114
       & \cellcolor{red}23.66 & \cellcolor{red}0.078 & \cellcolor{red}0.048 & \cellcolor{red}0.118 \\
\bottomrule
\end{tabular}}
\caption{Quantitative comparison on Neural3DV dataset (Part 2).}
\label{tab:group2}
\end{table*}

\begin{table*}[t]
\centering
\resizebox{\linewidth}{!}{
\begin{tabular}{lcccccccccccc}
\toprule
& \multicolumn{4}{c}{Actor1\_4} 
& \multicolumn{4}{c}{Actor2\_3} 
& \multicolumn{4}{c}{Actor5\_6} \\
\cmidrule(lr){2-5} \cmidrule(lr){6-9} \cmidrule(lr){10-13}
Method & PSNR & DSSIM$_1$ & DSSIM$_2$ & LPIPS 
& PSNR & DSSIM$_1$ & DSSIM$_2$ & LPIPS
& PSNR & DSSIM$_1$ & DSSIM$_2$ & LPIPS \\
\midrule
STGS~\cite{li2024spacetime}   & 15.50 & 0.314 & 0.186 & 0.690
       & 15.55 & 0.313 & 0.184 & 0.686
       & 15.44 & 0.316 & 0.187 & 0.682 \\
4DGS~\cite{yang2023real}   & \cellcolor{orange}23.24 & \cellcolor{orange}0.194 & \cellcolor{orange}0.119 & \cellcolor{orange}0.154
       & \cellcolor{orange}23.64 & \cellcolor{orange}0.186 & \cellcolor{orange}0.114 & \cellcolor{orange}0.154
       & \cellcolor{orange}23.71 & \cellcolor{orange}0.145 & \cellcolor{orange}0.087 & \cellcolor{orange}0.139 \\
Ex4DGS~\cite{lee2024fully} & 20.98 & 0.229 & 0.142 & 0.269
       & 22.02 & 0.236 & 0.145 & 0.283
       & 22.68 & 0.208 & 0.125 & 0.235 \\
\midrule
Ours   & \cellcolor{red}24.07 & \cellcolor{red}0.182 & \cellcolor{red}0.109 & \cellcolor{red}0.129
       & \cellcolor{red}24.39 & \cellcolor{red}0.175 & \cellcolor{red}0.104 & \cellcolor{red}0.124
       & \cellcolor{red}24.50 & \cellcolor{red}0.130 & \cellcolor{red}0.077 & \cellcolor{red}0.111 \\
\bottomrule
\end{tabular}}
\caption{Quantitative comparison on ENeRF-Outdoor dataset.}
\label{tab:group3}
\end{table*}

\end{document}